\documentclass[journal]{IEEEtran}
\usepackage[linesnumbered,ruled]{algorithm2e}
\usepackage{graphicx,amssymb,mathrsfs,amsmath,array,color,amsthm}
\usepackage{subfigure}
\usepackage{multirow}
\usepackage{booktabs}
\usepackage{soul}
\usepackage{bm}
\usepackage{mathtools}
\usepackage{bbm}

\usepackage{cite}
\usepackage[colorlinks,urlcolor=blue,driverfallback=dvipdfm]{hyperref}
\usepackage{pifont}
\newcommand{\cmark}{\ding{51}}
\newcommand{\xmark}{\ding{55}}

\DeclarePairedDelimiter{\floor}{\lfloor}{\rfloor}

\begin{document}
\title{SpectralFormer: Rethinking Hyperspectral Image Classification with Transformers}

\author{Danfeng Hong,~\IEEEmembership{Senior Member,~IEEE,}
        Zhu Han,~\IEEEmembership{Student Member,~IEEE,}
        Jing Yao,
        Lianru Gao,~\IEEEmembership{Senior Member,~IEEE,}
        Bing Zhang,~\IEEEmembership{Fellow,~IEEE,}
        Antonio Plaza,~\IEEEmembership{Fellow,~IEEE,}
        and Jocelyn Chanussot,~\IEEEmembership{Fellow,~IEEE}

\thanks{This work was supported in part by the National Natural Science Foundation of China under Grant 42030111 and Grant 41722108, and by the MIAI@Grenoble Alpes (ANR-19-P3IA-0003) and the AXA Research Fund. (\emph{Corresponding author: Lianru Gao})}
\thanks{D. Hong, J. Yao and L. Gao are with the Key Laboratory of Digital Earth Science, Aerospace Information Research Institute, Chinese Academy of Sciences, 100094 Beijing, China. (e-mail: hongdf@aircas.ac.cn; yaojing@aircas.ac.cn; gaolr@aircas.ac.cn)}
\thanks{Z. Han and B. Zhang are with the Key Laboratory of Digital Earth Science, Aerospace Information Research Institute, Chinese Academy of Sciences, Beijing 100094, China, and also with the College of Resources and Environment, University of Chinese Academy of Sciences, Beijing 100049, China. (e-mail: hanzhu19@mails.ucas.ac.cn; zb@radi.ac.cn)}
\thanks{A. Plaza is with the Hyperspectral Computing Laboratory, Department of Technology of Computers and Communications, Escuela Polit\'ecnica, University of Extremadura, 10003 C\'aceres, Spain. (e-mail: \mbox{aplaza@unex.es}).}
\thanks{J. Chanussot is with the Univ. Grenoble Alpes, INRIA, CNRS, Grenoble INP, LJK, F-38000 Grenoble, France, and also with the Aerospace Information Research Institute, Chinese Academy of Sciences, Beijing 100094, China. (e-mail: jocelyn@hi.is)}
}

\markboth{Submission to IEEE Transactions on Geoscience and Remote Sensing,~Vol.~XX, No.~XX, ~XXXX,~2021}
{Shell \MakeLowercase{\textit{et al.}}: Hyperspectral Image Classification with Transformers}

\maketitle
\begin{abstract}
Hyperspectral (HS) images are characterized by approximately contiguous spectral information, enabling the fine identification of materials by capturing subtle spectral discrepancies. Owing to their excellent locally contextual modeling ability, convolutional neural networks (CNNs) have been proven to be a powerful feature extractor in HS image classification. However, CNNs fail to mine and represent the sequence attributes of spectral signatures well due to the limitations of their inherent network backbone. To solve this issue, we rethink HS image classification from a sequential perspective with transformers, and propose a novel backbone network called \ul{SpectralFormer}. Beyond band-wise representations in classic transformers, SpectralFormer is capable of learning spectrally local sequence information from neighboring bands of HS images, yielding group-wise spectral embeddings. More significantly, to reduce the possibility of losing valuable information in the layer-wise propagation process, we devise a cross-layer skip connection to convey memory-like components from shallow to deep layers by adaptively learning to fuse ``soft'' residuals across layers. It is worth noting that the proposed SpectralFormer is a highly flexible backbone network, which can be applicable to both pixel- and patch-wise inputs. We evaluate the classification performance of the proposed SpectralFormer on three HS datasets by conducting extensive experiments, showing the superiority over classic transformers and achieving a significant improvement in comparison with state-of-the-art backbone networks. The codes of this work will be available at \url{https://github.com/danfenghong/IEEE_TGRS_SpectralFormer} for the sake of reproducibility.
\end{abstract}
\graphicspath{{figures/}}


\section{Introduction}
\IEEEPARstart{I}{n} hyperspectral (HS) imaging, hundreds of (narrow) wavelength bands are collected at each pixel across the complete electromagnetic spectrum, which enables the identification or detection of materials at a fine-grained level, particularly for those that have extremely similar spectral signatures in visual cues (e.g., RGB) \cite{hong2021interpretable}. This provides a great potential in a variety of high-level Earth observation (EO) missions, such as accurate land cover mapping, precision agriculture, target / object detection, urban planning, tree species classification, mineral exploration, and so on.

A general sequential process in a HS image classification system consists of image restoration (e.g., denoising, missing data recovery) \cite{wang2017hyperspectral,cao2018robust,wang2021,peng2021low}, dimensionality reduction \cite{hong2021joint,luo2020semisupervised}, spectral unmixing \cite{yao2019nonconvex,hong2019augmented,yuan2020improved,gao2021cycu,hong2021endmember}, and feature extraction \cite{hong2019learning,peng2018self,hong2020invariant,li2020ensemble}. Among them, feature extraction is a crucial step in HS image classification, which has received increasing attention by researchers. Over the past decade, a large number of advanced hand-crafted and subspace learning-based feature extraction approaches have been proposed for HS image classification \cite{rasti2020feature}. These methods are capable of performing well in small-sample classification problems. However, they tend to meet the performance bottleneck, when the training size gradually increases and the training set becomes more complex. The possible reason is due to the limited data fitting and representation ability of these traditional methods. Inspired by the great success of deep learning (DL), that is capable of finding out connotative, intrinsic, and potentially valuable knowledge from the vast amounts of pluralistic data \cite{lecun2015deep}, enormous efforts have been made in designing and adding advanced modules in networks to extract more diagnostic features from remote sensing data. For example, Zhao \textit{et al.} \cite{zhao2020joint} developed a joint classification framework using HS and light detection and ranging (LiDAR) data, which has been shown to be excellent at extracting features from multisource RS data. Zhang \textit{et al.} \cite{zhang2018feature} designed an extraordinary patch-to-patch convolutional neural network (CNN), which has yielded significantly better results than other techniques.

\begin{figure*}[!t]
	  \centering
			\includegraphics[width=1\textwidth]{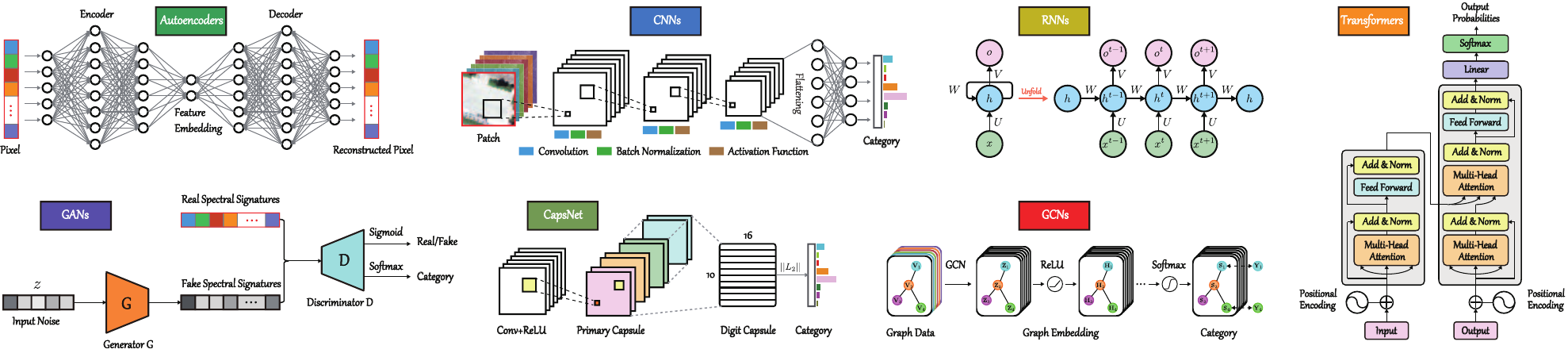}
        \caption{An overview of currently well-recognized backbone networks for the HS image classification task, such as autoencoders \cite{chen2014deep}, CNNs \cite{chen2016deep}, RNNs \cite{hang2019cascaded}, GANs \cite{zhu2018generative}, CapsNet \cite{paoletti2018capsule}, GCNs \cite{hong2021graph}, and Transformers \cite{vaswani2017attention}.}
\label{fig:overview}
\end{figure*}

In recent years, many well-recognized backbone networks have been widely and successfully applied in the HS image classification task \cite{li2019deep}, such as autoencoders (AEs), CNNs, recurrent neural networks (RNNs), generative adversarial networks (GANs), capsule networks (CapsNet), graph convolutional networks (GCNs). Chen \textit{et al.} \cite{chen2014deep} stacked multiple autoencoder networks to extract deep feature representations from dimension-reduced HS images -- generated by principal component analysis (PCA) \cite{abdi2010principal} -- with the application to HS image classification. Chen \textit{et al.} \cite{chen2016deep} employed CNNs instead of stacked AEs to semantically extract spatial-spectral features by considering local contextual information of HS images, achieving higher classification accuracies. Hang \textit{et al.} \cite{hang2019cascaded} designed a cascaded RNN for HS image classification by taking advantage of RNNs that can model the sequentiality to represent the relations of neighboring spectral bands effectively. In \cite{zhu2018generative}, GANs were improved, making them applicable to the HS image classification task, with the input of three PCA components and random noises. Paoletti \textit{et al.} \cite{paoletti2018capsule} extended a CNN-based model by defining a novel spatial-spectral capsule unit, yielding a high-performance classification framework of HS images and meanwhile reducing the complexity of the network design. Hong \textit{et al.} \cite{hong2021graph} made a comprehensive comparison between CNNs and GCNs on the classification of HS images both qualitatively and quantitatively, and proposed a mini-batch GCN (miniGCNs), providing a feasible solution for addressing the issue of large graphs in GCNs, for state-of-the-art HS image classification.

Although these backbone networks and their variants have been able to obtain promising classification results, their ability in characterizing spectral series information (particularly in capturing subtle spectral discrepancies along spectral dimension) remains inadequate. Fig. \ref{fig:overview} gives an overview illustration of these state-of-the-art backbone networks in the HS image classification task. The specific limitations can be roughly summarized as follows.
\begin{itemize}
    \item As a mainstream backbone architecture, CNNs have shown their powerful ability in extracting spatially structural information and locally contextual information from HS images. Nevertheless, \textit{on the one hand,} CNNs can hardly be capable of capturing the sequence attributes well, particularly middle- and long-term dependencies. This unavoidably meets the performance bottleneck in the HS image classification task, especially when the to-be-classified categories are of a great variety and extremely similar in spectral signatures. \textit{On the other hand,} CNNs are overly concerned with spatial content information, which spectrally distorts sequential information in the learned features. This, to great extent, enables more difficulties of mining diagnostic spectral attributes.  
    \item Unlike CNNs, RNNs are designed for sequence data, which accumulatively learns spectral features band-by-band from HS images in an orderly fashion. Such a mode extremely depends on the order of spectral bands and tends to generate gradient vanishing, thereby being hard to learn long-term dependencies \cite{bengio1994learning}. This might further lead to the difficulty of capturing spectrally salient changes in time series. More importantly, there are usually tons of HS samples (or pixels) available in real HS image scenes, yet RNNs fail to train the models in parallel, limiting the classification performance in practical applications. 
    \item For other backbone networks, i.e., GANs, CapsNet, GCNs, despite their respective advantages in learning spectral representations (e.g., robustness, equivalence, long-range relevance between samples), one thing in common is that almost all of them could be inherently incompetent for modeling sequential information effectively. That is, the utilization of spectral information is insufficient (being a critical bottleneck in fine land cover classification or mapping using HS data).
\end{itemize}

Being aimed at the aforementioned limitations, we rethink the HS image classification process from a sequence data perspective with current state-of-the-art transformers \cite{vaswani2017attention}. Totally different from CNNs and RNNs, transformers are (at present) one of the cutting-edge backbone networks, owing to the use of self-attention techniques, which are well-designed for the sake of processing and analyzing sequential (or time series) data more effectively. This will provide a good fit for HS data processing and analysis, e.g., HS image classification. It is well-known that the self-attention block in transformers enables to capture globally sequential information by the means of positional encoding \cite{ke2020rethinking}. However, there also exist some drawbacks in transformers that hinder its performance to be further improved. For example, 
\begin{itemize}
    \item[1)] although transformers perform outstandingly in solving the problem of long-term dependencies of spectral signatures, they lose the power of capturing locally contextual or semantic characteristics (\textit{cf.} CNNs or RNNs);
    \item[2)] as mentioned in \cite{dong2021attention}, the skip connection plays a crucial role in transformers. This might be explained well by either using ``residuals'' to make the gradients better propagated or enhancing ``memories'' to reduce the forgetting or loss of key information. But unfortunately, the simple additive skip connection operation only occurs within each transformer block, weakening the connectivity across different layers or blocks. 
\end{itemize}

For these reasons, we aim to develop a novel transformers-based network architecture, SpectralFormer for short, enabling the high-performance HS image classification task. SpectralFormer provides point-to-point solutions corresponding to the above two drawbacks. More specifically, SpectralFormer is capable of learning locally spectral representations from multiple neighboring bands rather than single band (in original transformers) in each encoded position, e.g., \textit{group-wise versus band-wise embeddings}. Furthermore, a cross-layer skip connection is designed in SpectralFormer to progressively convey memory-like components from shallow to deep layers by adaptively learning to fuse their ``soft'' residuals. The main contributions of this paper can be summarized as follows.
\begin{itemize}
    \item We revisit the HS image classification problem from a sequential perspective and propose a new transformers-based backbone network, called SpectralFormer, in order to substitute for CNNs or RNNs-based architectures. To the best of our knowledge, this is the first time that transformers (without any pre-processing operation, e.g., feature extraction using convolution and recurrent units or other transformation techniques) are purely applied to the HS image classification task.
    \item We devise two simple but effective modules in SpectralFormer, i.e., group-wise spectral embedding (GSE) and cross-layer adaptive fusion (CAF), to learn locally detailed spectral representations and convey memory-like components from shallow to deep layers, respectively.
    \item We qualitatively and quantitatively evaluate the classification performance of the proposed SpectralFormer on three representative HS datasets, i.e., Indian Pines, Pavia University, and University of Houston, with extensive ablation studies. The experimental results demonstrate a significant superiority over classic transformers (with an increase of approximately 10\% OA) and other state-of-the-art backbone networks (at least 2\% OA improvement).
\end{itemize}

The remaining of the paper is organized as follows. Section II first reviews classic transformers-related literature and then details the proposed SpectralFormer with two well-designed modules for HS image classification. Extensive experiments are conducted with ablation studies and discussions in Section III. Section IV draws comprehensive conclusions and a brief outlook on future possible research directions.

\section{SpectralFormer}
In this section, we start to review some preliminaries of classic transformers. On this basis, we propose the SpectralFormer with two well-designed modules, i.e., GSE and CAF, making it more applicable to the HS image classification task. Finally, we also investigate the ability that the proposed SpectralFormer models the spatially contextual information with the input of image patches.

\subsection{A Brief Review of Transformers}
As is known to all, transformers \cite{vaswani2017attention} have cut a conspicuous figure in processing sequence-to-sequence problems in natural language processing (NLP), e.g., machine translation. Since they abandon the sequence dependence characteristic in RNNs and alternatively introduce a brand-new self-attention mechanism. This enables the global information (long-term dependencies) capture for the units at any position, which greatly promotes the development of time series data processing models. Even not only limiting in NLP, image processing and computer vision fields have also begun exploring the transformer architecture. Very recently, the vision transformer (ViT) \cite{dosovitskiy2020image} seems to have achieved or approached CNNs-based state-of-the-art effects on various vision domain tasks, providing new insight, inspiration, and creative space on vision-related tasks. 

\begin{figure}[!t]
	  \centering
		\subfigure[Self-attention]{
			\includegraphics[width=0.12\textwidth]{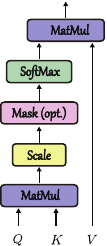}
			\label{fig:attention_1}
		}\;\;\;\;\;
		\subfigure[Multi-head attention]{
			\includegraphics[width=0.28\textwidth]{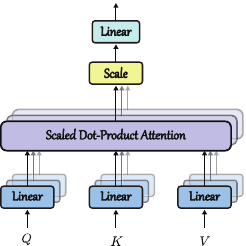}
			\label{fig:attention_2}
		}
		\caption{An illustration of the attention mechanism in transformers. (a) Self-attention module; (b) Multi-head attention.}
\label{fig:attention}
\end{figure}

\begin{figure*}[!t]
	  \centering
			\includegraphics[width=1\textwidth]{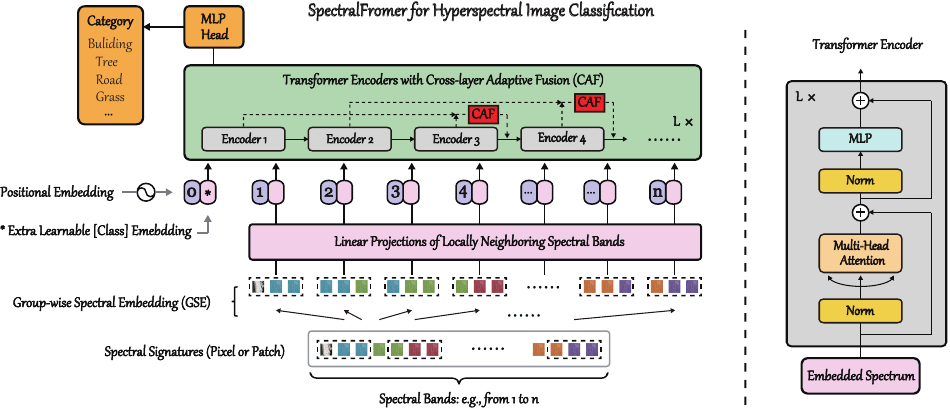}
        \caption{An overview illustration of the proposed SpectralFormer network for the HS image classification task. SpectralFormer consists of two well-designed modules, i.e., GSE and CAF, making it (the transformers-based backbone) better applicable to HS images.}
\label{fig:workflow}
\end{figure*}

The success of transformers, to a great extent, depends on the use of multi-head attention, where multiple self-attention (SA) \cite{wang2018non} layers are stacked and integrated. As the name suggests, SA mechanism is better at capturing the internal correlation of data or features, thereby reducing the dependence on external information. Fig. \ref{fig:attention_1} illustrates the process of the SA module in transformers. More specifically, the SA layer can be performed according to the following six steps:

{\bf Step 1.} Input the sequence data $\bm{x}$ with the length of $m$, where $\bm{x}_i, i = 1, ..., m$ denotes either a scalar or a vector.

{\bf Step 2.} The feature embedding, denoted as $\bm{a}_{i}$, is obtained on each scalar or vector $\bm{x}_{i}$ by a shared matrix $\bm{W}$.

{\bf Step 3.} Each embedding multiplies by three different transformation matrices $\bm{W}_{q}$, $\bm{W}_{k}$, $\bm{W}_{v}$, respectively, yielding three vectors, i.e., Query ($Q=[\bm{q}_{1},...\bm{q}_{m}]$), Key ($K=[\bm{k}_{1},...\bm{k}_{m}]$), Value ($V=[\bm{v}_{1},...,\bm{v}_{m}]$). 

{\bf Step 4.} Compute the attention score $\bm{s}$ between each $Q$ vector and each $K$ vector in the form of inner product, e.g., $\bm{q}_{i}\cdot \bm{k}_{j}$, and to stabilize the gradients, the scaled score is obtained by normalization, i.e., $\bm{s}_{i,j}=\bm{q}_{i}\cdot \bm{k}_{j}/\sqrt{d}$, where $d$ is the dimension of $\bm{q}_{i}$ or $\bm{k}_{j}$.

{\bf Step 5.} The Softmax activation function is performed on $\bm{s}$; we then have in the position-1 for instance: $\bm{\hat{s}}_{1,i}=e^{\bm{s}_{1,i}}/\sum_{j}e^{\bm{s}_{1,j}}$.

{\bf Step 6.} Generate  attention representations $\bm{z}=[\bm{z}_{1},...,\bm{z}_{m}]$, e.g., $\bm{z}_{1}=\sum_{i}\bm{\hat{s}}_{1,i}\bm{v}_{i}$.

To sum up, the SA layer can be integrally formulated as follows
\begin{equation}
\label{eq1}
\begin{aligned}
       \bm{z}={\rm Attention}(Q, K, V)={\rm softmax}(\frac{QK^{\top}}{\sqrt{d}})V.
\end{aligned}
\end{equation} 
With Eq. (\ref{eq1}), multiple different SA layers can be assembled to be a multi-head attention, as shown in Fig. \ref{fig:attention_2}. In detail, we first obtain multiple attention representations (e.g., $h=8$), denoted as $z^{h}, h = 1,...,8$, and concatenate them to be a larger feature matrix. A linear transformation matrix (e.g., $\bm{W}_{o}$) is finally used to make the feature dimension identical to the input data.

It should be noted, however, that there is no position information in the SA layer, which fails to make use of sequence information. For this reason, the position information is encoded into the feature embedding. The embedding with time signal can be thus formulated as $\bm{a}_{i} + \bm{e}_{i}$, where $\bm{e}_{i}$ denotes a unique positional vector given manually.

\begin{table*}[!t]
\centering
\caption{Definition of notations used in the proposed SpectralFormer}
\begin{tabular}{c||c|c|c}
\toprule[1.5pt]
Notation & Definition & Type & Size\\
\hline \hline
$m$ & the number of spectral bands & scalar & $1\times 1$ \\
$\bm{x}$ & spectral signature of a HS pixel & vector & $1\times m$\\
$x_{i}$ & reflectance or radiance value in the $i$-th band location & scalar & $1\times 1$ \\
$d$ & the dimension of the feature embeddings & scalar & $1\times 1$ \\
$w$ & the linear transformation w.r.t a pixel (sample) & vector & $d\times 1$ \\
$\bm{A}$ & the feature embeddings & matrix & $d\times m$ \\
$n$ & the number of considered HS pixels & scalar & $1\times 1$ \\
$\bm{W}$ & the linear transformations w.r.t. $n$ pixels (samples) & matrix & $d\times n$ \\
$\bm{X}$ & spectral signatures of $n$ pixels & matrix & $n\times m$\\
$g(\cdot)$ & the overlapping grouping operation w.r.t. the variable $(\cdot)$ & -- & -- \\
$\bm{\dot{A}}$ & the feature embeddings of GSE & matrix & $d\times m$ \\
$\floor*{\cdot}$ & the round operation & -- & -- \\
$\bm{z}^{(l)}$ & the feature representations in the $(l)$-th layer & vector & $1\times d_{z}$ \\
$\bm{\ddot{w}}$ & the learnable network parameter for adaptive fusion & vector & $1\times 2$ \\
$\bm{\hat{z}}^{(l)}$ & the fused representations in the $(l)$-th layer with the CAF & vector & $1\times d_{z}$ \\
\bottomrule[1.5pt]
\end{tabular}
\label{table:notations}
\end{table*}

\subsection{Overview of SpectralFormer}
We aim at developing a novel and generic ViT-based baseline network (i.e., SpectralFormer) with a focus on the spectrometric characteristics, making it well-applicable to the highly accurate and fine classification of HS images. To this end, we devise two key modules, i.e., GSE and CAF, and integrate them into the transformer framework, in order to improve the detail-capturing capacity of subtle spectral discrepancies and enhance the information transitivity (or connectivity) between layers (i.e., reduce the information loss with the gradual deepening of layers), respectively. Moreover, the proposed SpectralFormer is not only applied to the pixel-wise HS image classification, but also extensible to the spatial-spectral classification with the batch-wise input, yielding the spatial-spectral SpectralFormer version. Fig. \ref{fig:workflow} illustrates an overview of the proposed SpectralFormer in the HS image classification task, while Table \ref{table:notations} details the definition of notations used in the proposed SpectralFormer.

\subsection{Group-wise Spectral Embedding (GSE)}
\begin{figure}[!t]
	  \centering
		\subfigure[Band-wise Spectral Embedding (BSE)]{
			\includegraphics[width=0.48\textwidth]{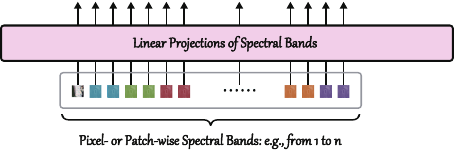}
			\label{fig:band-wise}
		}
		\subfigure[Group-wise Spectral Embedding (GSE)]{
			\includegraphics[width=0.48\textwidth]{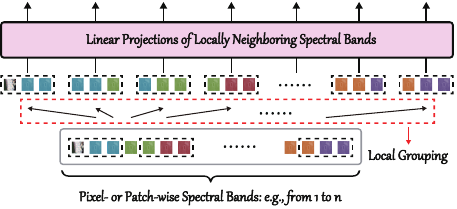}
			\label{fig:group-wise}
		}
		\caption{An illustrative comparison of band-wise and group-wise spectral embeddings (BSE \textit{versus} GSE) in transformers.}
\label{fig:BandvsGroup}
\end{figure}

Unlike the discrete sequentiality in classic transformers or ViT (e.g., image patches), hundreds or thousands of spectral channels in HS images are densely sampled from the electromagnetic spectrum at a subtle interval (e.g., 10nm), yielding approximately continuous spectral signatures. The spectral information in different position reflects different absorption characteristics corresponding to different wavelength. This to a great extent shows the physical properties of the current material. Capturing locally detailed absorption (or changes) of such spectral signatures is a crucial factor to accurately and finely classify the materials lying in the HS scene. For this purpose, we propose to learn group-wise spectral embeddings, i.e., GSE, instead of band-wise input and representations. 
Given a spectral signature (a pixel in the HS image) $\bm{x}=[x_1,x_2,\ldots,x_m]\in \mathbb{R}^{1\times m}$, the feature embeddings $\bm{a}$ obtained by classic transformers are formulated by
\begin{equation}
\label{eq2}
\begin{aligned}
       \bm{A}=\bm{w}\bm{x},
\end{aligned}
\end{equation} 
where $\bm{w}\in \mathbb{R}^{d\times 1}$ denotes the linear transformation that is equivalently used for all bands in spectral signatures and $\bm{A}\in \mathbb{R}^{d\times m}$ collects the output features. Whereas the proposed GSE learns the feature embeddings from locally spectral profiles (or neighboring bands). Thus, the GSE can be modeled as
\begin{equation}
\label{eq3}
\begin{aligned}
       \bm{\dot{A}}=\bm{W}\bm{X}=\bm{W}g(\bm{x}),
\end{aligned}
\end{equation} 
where $\bm{W}\in \mathbb{R}^{d\times n}$ and $\bm{X}\in \mathbb{R}^{n\times m}$ correspond to the grouped representations with respect to the variables $\bm{w}$ and $\bm{x}$, respectively, $n$ represents the number of neighboring bands. The variable $\bm{W}$ can be simply seen as one layer of network, which can be optimized by updating the whole network. The function $g(\cdot)$ denotes the overlapping grouping operation in regard to the variable $\bm{x}$, i.e.,
\begin{equation}
\label{eq4}
\begin{aligned}
    \bm{X}=g(\bm{x})=[\bm{x}_{1},...,\bm{x}_{q},...,\bm{x}_{m}],
\end{aligned}
\end{equation}
where $\bm{x}^{q}=[x_{q-\floor*{\frac{n}{2}}},...,x_{q},...,x_{q+\floor*{\frac{n}{2}}}]^\top\in \mathbb{R}^{n\times 1}$, $\floor*\bullet$ denotes the round operation. Fig. \ref{fig:BandvsGroup} illustrates the differences between band-wise and group-wise spectral embeddings in transformers-based backbone networks, i.e., BSE \textit{versus} GSE.

\subsection{Cross-layer Adaptive Fusion (CAF)}
\begin{figure}[!t]
	  \centering
			\includegraphics[width=0.48\textwidth]{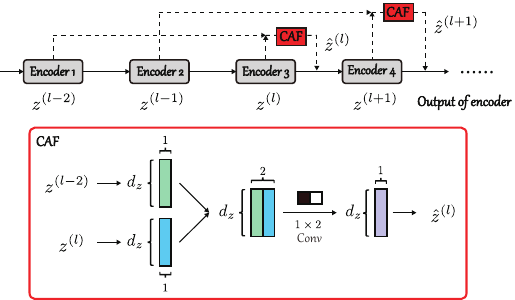}
        \caption{A diagram of cross-layer adaptive fusion (CAF) module in the proposed SpectralFormer.}
\label{fig:CAF}
\end{figure}

The skip connection (SC) mechanism has been proven to be an effective strategy in deep networks, which can enhance information exchange between layers and reduce information loss in the network learning process. The use of SC has recently gained tremendous success in image recognition and segmentation, e.g., short SC for ResNet \cite{he2016deep} and long SC for U-Net \cite{ronneberger2015u}. It should be noted, however, that the information ``memory'' ability of the short SC remains limited, while the long SC tends to yield insufficient fusion due to a big gap between high- and low-level features. This is also a key problem existed in transformers, which will pose a new challenge to the architecture design of transformers. To this end, we devise a middle-range SC in SpectralFormer to adaptively learn cross-layer feature fusion (i.e., CAF, see Fig. \ref{fig:CAF}).

Let $\bm{z}^{(l-2)}\in \mathbb{R}^{1\times d_{z}}$ and $\bm{z}^{(l)}\in \mathbb{R}^{1\times d_{z}}$ be the outputs (or representations) in the $(l-2)$-th and $(l)$-th layers, respectively, CAF can be then expressed by
\begin{equation}
\label{eq5}
\begin{aligned}
       \bm{\hat{z}}^{(l)}\leftarrow \bm{\ddot{w}}\left[
           	 \begin{matrix}
                \bm{z}^{(l)}\\
    	        \bm{z}^{(l-2)}
         \end{matrix}\right],
\end{aligned}
\end{equation} 
where $\bm{\hat{z}}^{(l)}$ denotes the fused representations in the $(l)$-th layer with the proposed CAF, and $\bm{\ddot{w}}\in \mathbb{R}^{1\times 2}$ is the learnable network parameter for adaptive fusion. It should be noted that our CAF only skips one encoder, e.g., from $\bm{z}^{(l-2)}$ (the output of Encoder 1) to $\bm{z}^{(l)}$ (the output of Encoder 3) in Fig. \ref{fig:CAF}. The reasons for this setting are two-fold. On the one hand, there is a big semantic gap between low- and deep-level features obtained from the shallow and deep layers, respectively. If the relatively long SC, e.g., two, three, and even more encoders, is used, this then might lead to an insufficient fusion and potential information loss. On the other hand, the HS image classification can be often regarded as a small-sample problem, due to the limited available training (need to be labeled manually) samples. A ``small'' or ``shallow'' backbone network, e.g., 4 or 5 layers, may already be a good fit for the HS image classification task. As a result, this, to some extent, can explain why we propose to skip only one encoder in the CAF module (since a 4 or 5-layered shallow network is small, which is not capable of adding multiple CAF modules).

\subsection{Spatial-Spectral SpectralFormer}
Beyond the pixel-wise HS image classification, we similarly investigate the patch-wise input (inspired by CNNs), yielding the spatial-spectral SpectralFormer version, i.e., patch-wise SpectralFormer. Different from CNNs, that directly input a 3-D patch cube, we unfold the 2-D patch of each band to the corresponding 1-D vector representations. Given a 3-D cube $\mathcal{X}\in \mathbb{R}^{m\times w\times h}$ ($w$ and $h$ are the width and length of the patch), which can be unfolded along with the spatial direction, we then have
\begin{equation}
\label{eq6}
\begin{aligned}
       \bm{\hat{X}}=[\bm{\vec{x}}_{1},...,\bm{\vec{x}}_{i},...,\bm{\vec{x}}_{m}],
\end{aligned}
\end{equation} 
where $\bm{\vec{x}}_{i}\in \mathbb{R}^{wh\times 1}$ denotes the unfolded patch for the $i$-th band. Such an input way can to a great extent preserve the spectrally sequential information in network learning and meanwhile consider spatially contextual information.

\section{Experiments}

In this section, three well-known HS datasets are firstly described, the implementation details and compared state-of-the-art methods are then introduced. Finally, extensive experiments are conducted with ablation analysis to assess the HS image classification performance of the proposed SpectralFormer both quantitatively and qualitatively. 

\subsection{Data Description}
\subsubsection{Indian Pines Data} The first HS data was collected in 1992 by using the Airborne Visible/Infrared Imaging Spectrometer (AVIRIS) sensor over North-Western Indiana, USA. The HS image consists of $145\times 145$ pixels at a ground sampling distance (GSD) of $20$m, and $220$ spectral bands covering the wavelength range of $400$nm to $2500$nm with a $10$m spectral resolution. After removing $20$ noisy and water absorption bands, $200$ spectral bands are retained, i.e., 1-103, 109-149, 164-219. There are 16 mainly-investigated categories in this studied scene. The class name and the number of samples used for training and testing in the classification task are listed in Table \ref{Table:Indian}, while the spatial distribution of training and testing sets are also given in Fig. \ref{fig:CM_IP} to reproduce the classification results.

\subsubsection{Pavia University Data} The second HS scene was acquired by the Reflective Optics System Imaging Spectrometer (ROSIS) sensor over Pavia University and its surroundings, Pavia, Italy. The sensor can capture $103$ spectral bands ranging from $430$nm to $860$nm, and the image consists of $610\times 340$ pixels at a GSD of $1.3$m. This scene includes $9$ land cover classes, where the fixed number of training and testing samples are detailed in Table \ref{Table:Pavia} and spatially visualized in Fig. \ref{fig:CM_PU}.

\subsubsection{Houston2013 Data} The third dataset was acquired by the ITRES CASI-1500 sensor over the campus of the University of Houston and its neighboring rural areas, Texas, USA, which, as a benchmark dataset, has been widely used for evaluating the performance of land cover classification \cite{hong2021more}. The HS cube comprises of  $349\times 1905$ pixels with $144$ wavelength bands in the range of $364$nm to $1046$nm at $10$nm intervals. It is worth noting, however, that the Houston2013 dataset we used is a cloud-free version, which is processed to recover the missing data or remove occlusions by generating illumination-related threshold maps\footnote{The data were provided by Prof. N. Yokoya from the University of Tokyo and RIKEN AIP.}. Table \ref{Table:H2013} lists $15$ challenging land-cover and land-use categories as well as the corresponding sample number of training and testing sets. Similarly, the visualization results with respect to the false-color HS image and the released training and testing samples provided by the 2013 IEEE GRSS data fusion contest\footnote{http://www.grss-ieee.org/community/technical-committees/data-fusion/2013-ieee-grss-data-fusion-contest/} are given in Fig. \ref{fig:CM_HH}.

\begin{table}[!t]
\centering
\caption{Land-cover classes of the Indian Pines dataset, with the standard training and testing sets for each class.}
\begin{tabular}{c||ccc}
\toprule[1.5pt]
Class No.&Class Name&Training&Testing\\
\hline \hline 1&Corn Notill&50&1384\\
 2&Corn Mintill&50&784\\
 3&Corn&50&184\\
 4&Grass Pasture&50&447\\
 5&Grass Trees&50&697\\
 6&Hay Windrowed&50&439\\
 7&Soybean Notill&50&918\\
 8&Soybean Mintill&50&2418\\
 9&Soybean Clean&50&564\\
 10&Wheat&50&162\\
 11&Woods&50&1244\\
 12&Buildings Grass Trees Drives&50&330\\
 13&Stone Steel Towers&50&45\\
 14&Alfalfa&15&39\\
 15&Grass Pasture Mowed&15&11\\
 16&Oats&15&5\\
\hline \hline &Total&695&9671\\
\bottomrule[1.5pt]
\end{tabular}
\label{Table:Indian}
\end{table}

\begin{table}[!t]
\centering
\caption{Land-cover classes of the Pavia University dataset, with the standard training and testing sets for each class.}
\begin{tabular}{c||ccc}
\toprule[1.5pt]
Class No.&Class Name&Training&Testing\\
\hline \hline 1&Asphalt&548&6304\\
 2&Meadows&540&18146\\
 3&Gravel&392&1815\\
 4&Trees&524&2912\\
 5&Metal Sheets&265&1113\\
 6&Bare Soil&532&4572\\
 7&Bitumen&375&981\\
 8&Bricks&514&3364\\
 9&Shadows&231&795\\
\hline \hline &Total&3921&40002\\
\bottomrule[1.5pt]
\end{tabular}
\label{Table:Pavia}
\end{table}

\begin{table}[!t]
\centering
\caption{Land-cover classes of the Houston2013 dataset, with the standard training and testing sets for each class.}
\begin{tabular}{c||ccc}
\toprule[1.5pt]
Class No.&Class Name&Training&Testing\\
\hline \hline 1&Healthy Grass&198&1053\\
 2&Stressed Grass&190&1064\\
 3&Synthetic Grass&192&505\\
 4&Tree&188&1056\\
 5&Soil&186&1056\\
 6&Water&182&143\\
 7&Residential&196&1072\\
 8&Commercial&191&1053\\
 9&Road&193&1059\\
 10&Highway&191&1036\\
 11&Railway&181&1054\\
 12&Parking Lot1&192&1041\\
 13&Parking Lot2&184&285\\
 14&Tennis Court&181&247\\
 15&Running Track&187&473\\
\hline \hline &Total&2832&12197\\
\bottomrule[1.5pt]
\end{tabular}
\label{Table:H2013}
\end{table}

\subsection{Experimental Setup}
\subsubsection{Evaluation Metrics}
We evaluate the classification performance of each model quantitatively in terms of three commonly-used indices, i.e., \textit{Overall Accuracy (OA)}, \textit{Average Accuracy (AA)}, \textit{Kappa Coefficient ($\kappa$)}. Moreover, the classification maps obtained by different models are visualized to make a qualitative comparison.

\subsubsection{Comparison with State-of-the-art Backbone Networks}
Several representative baselines and backbone networks are selected for the following comparison experiments. They are K-nearest neighbor (KNN), support vector machine (SVM), random forest (RF), 1-D CNN \cite{rasti2020feature}, 2-D CNN \cite{chen2016deep}, RNN \cite{hang2019cascaded}, miniGCN \cite{hong2021graph}, transformers \cite{vaswani2017attention}, and the proposed SpectralFormer. The parameter configurations of these compared methods are detailed below:
\begin{itemize}
    \item For the KNN, the number of nearest neighbors ($K$) is an important factor, which greatly impacts the classification performance. We set it to 10.
    \item For the RF, $200$ decision trees are used in our experiments.
    \item For the SVM, the libsvm toolbox\footnote{https://www.csie.ntu.edu.tw/$\sim$cjlin/libsvm/} is selected for the implementation of the HS image classification task. The SVM is performed by using the radial basis function (RBF) kernel. In RBF, two hyperparameters $\sigma$ and $\lambda$ can be optimally determined by five-fold cross validation on the training set in the range of $\sigma=[2^{-3},2^{-2},\dots,2^{4}]$ and $\lambda=[10^{-2},10^{-1},\dots,10^{4}]$, respectively.
    \item For the 1-D CNN, one convolutional block is defined as the basic network unit, including a set of 1-D convolutional filters with the output size of $128$, a batch normalization (BN) layer, and a ReLU activation function. A softmax function is finally added on the top layer of the 1-D CNN.
    \item The 2-D CNN architecture has three 2-D convolutional blocks and a softmax layer. Similar to the 1-D CNN, each convolutional block of 2-D CNN consists of a 2-D conventional layer, a BN layer, a max-pooling layer, and a ReLU activation function. Moreover, the spatially and spectrally receptive fields in each 2-D convolutional layer are $3\times 3 \times 32$, $3\times 3 \times 64$, and $1\times 1 \times 128$, respectively.
    \item For the RNN, there are two recurrent layers with the gated recurrent unit (GRU). Each of them has $128$ neuron units. The used codes are openly available from \url{https://github.com/danfenghong/HyFTech}.
    \item For the miniGCN, the network block successively contains a BN layer, a graph convolutional layer with $128$ neuron units, and a ReLU layer. Note that the adjacency matrix in GCN can be generated using a KNN-based graph (the number of $K$ is the same with the KNN classifier, i.e., $K=10$). The miniGCN and 1-D CNN share the same network architecture (for a fair comparison). As the name suggests, the miniGCN can be trained in a mini-batch fashion. We refer to \cite{hong2021graph} for more details and the codes\footnote{\url{https://github.com/danfenghong/IEEE_TGRS_GCN}} for the sake of reproducibility.
    \item For the transformers\footnote{The codes of transformers will be provided by authors, making it applicable to the HS image classification task.}, we follow the ViT network architecture \cite{dosovitskiy2020image}, i.e., only including transformer encoders. In detail, five encoder blocks are used in the ViT-based network for HS image classification.
    \item For the proposed SpectralFormer, we adopt the same backbone architecture as the above transformers for a fair comparison. More specifically, the embedded spectrum with $64$ units are fed into $5$ cascaded transformer encoder blocks for HS image classification. Each encoder block consists of a four-head SA layer, a MLP with $8$ hidden dimensions, and a GELU \cite{hendrycks2016gaussian} nonlinear activation layer. The dropout layer is employed after encoding positional embeddings and in MLPs for inhibiting $10\%$ neurons. Considering the fact that the parameter size is evidently increased from the pixel-wise SpectralFormer to the patch-wise one (the patch size is empirically set as $7\times 7$), we additionally employ an $\ell_2$ weight decay regularization \cite{loshchilov2017decoupled} parameterized by $5e-3$ to prevent a potential overfitting risk for the latter.
\end{itemize}

\subsubsection{Implementation Details}
Our proposed SpectralFormer was implemented on the PyTorch platform using a workstation with i7-6850K CPU, 128GB RAM, and an NVIDIA GTX 1080Ti 11GB GPU. We adopt the Adam optimizer \cite{kingma2014adam} with a mini-batch size of $64$. The learning rate is initialized with $5e-4$ and decayed by multiplying a factor of $0.9$ after each one-tenth of total epochs. We roughly set the epochs on the three datasets to $1000$\footnote{The epochs might be slightly adjusted for different algorithms and different datasets, which will be specifically provided in our codes.}. It is worth noting, however, that we found in practice our SpectralFormer with the CAF module is capable of embracing an evident efficiency improvement by convergence using much less epochs, i.e., $300$ epochs for the Indian Pines dataset, and $600$ epochs for the other two datasets. 

\begin{table}[!t]
    \centering
    \caption{Ablation analysis of the proposed SpectralFormer with a combination of different modules on the Indian Pines dataset.}
    \resizebox{0.48\textwidth}{!}{
    \begin{tabular}{c|c||cc||ccc}
        \toprule[1.5pt]
        \multirow{2}{*}{Method} & \multirow{2}{*}{Input} & \multicolumn{2}{c||}{Module} & \multicolumn{3}{c}{Metric} \\
        \cline{3-7} & & GSE & CAF & OA (\%) & AA (\%) & $\kappa$ \\
        \hline\hline
        Transformers (ViT) & pixel-wise & \xmark & \xmark & 71.86 & 78.97 & 0.6804\\
        SpectralFormer & pixel-wise & \cmark & \xmark & 75.05 & 82.91 & 0.7175\\
        SpectralFormer & pixel-wise & \xmark & \cmark & 74.37 & 80.87 & 0.7084\\
        SpectralFormer & pixel-wise & \cmark & \cmark & 78.55 & 84.68 & 0.7554\\
        \hline \hline
        SpectralFormer & patch-wise & \cmark & \cmark & \bf 81.76 & \bf 87.81 & \bf 0.7919\\
        \bottomrule[1.5pt]
    \end{tabular}
    }
    \label{tab:ablation}
\end{table}

\begin{table}[!t]
    \centering
    \caption{Sensitivity analysis of the number of neighboring bands in SpectralFormer (only GSE and GSE + CAF) in terms of \textit{OA}, \textit{AA}, and $\kappa$ on the Indian Pines dataset.}
    \resizebox{0.48\textwidth}{!}{
    \begin{tabular}{c|c||cccccc}
        \toprule[1.5pt]
         \multirow{2}{*}{SpectralFormer} & \multirow{2}{*}{Metric} & \multicolumn{6}{c}{The number of neighboring bands} \\
        \cline{3-8} & & 1 & 3 & 5 & 7 & 9 & 11 \\
        \hline\hline
        \multirow{3}{*}{Only GSE} & OA (\%) & 71.86 & 74.21 & 72.77 & \bf 75.05 & 74.13 & 74.11\\
        & AA (\%) & 78.97 & 83.68 & 82.02 & \bf 82.91 & 83.54 & 82.48\\
        & $\kappa$ & 0.6804 & 0.7088 & 0.6918 & \bf 0.7175 & 0.7074 & 0.7067\\
        \hline
        \multirow{3}{*}{GSE + CAF} & OA (\%) & 74.37 & \bf 78.55 & 77.11 & 77.41 & 76.46 & 77.65\\
        & AA (\%) & 80.87 & \bf 84.68 & 84.55 & 84.20 & 83.45 & 83.85\\
        & $\kappa$ & 0.7084 & \bf 0.7554 & 0.7394 & 0.7426 & 0.7319 & 0.7457\\
        \bottomrule[1.5pt]
    \end{tabular}
    }
    \label{tab:Sensitivity}
\end{table}

\begin{table}[!t]
    \centering
    \caption{Classification performance analysis between different SCs in transformers on the Indian Pines dataset.}
    \resizebox{0.48\textwidth}{!}{
    \begin{tabular}{c|c|c||ccc}
        \toprule[1.5pt]
         \multirow{2}{*}{Method} & \multirow{2}{*}{Input} & \multirow{2}{*}{SC} & \multicolumn{3}{c}{Metric} \\
        \cline{4-6} & & & OA (\%) & AA (\%) & $\kappa$ \\
        \hline\hline
        Transformers (ViT) & \multirow{3}{*}{pixel-wise} & short-range & 71.86 & 78.97 & 0.6804\\
        Transformers (ViT) & & long-range & 67.70 & 73.12 & 0.6315\\
        Transformers (ViT) & & CAF & \bf 74.37 & \bf 80.87 & \bf 0.7084\\
        \bottomrule[1.5pt]
    \end{tabular}
    }
    \label{tab:skipconnection}
\end{table}

\subsubsection{Computational Complexity Analysis}
For a given HS image with spectral length $m$, the per-layer computational complexity of the proposed SpectralFormer is mainly dominated by self-attention and multi-head operations that require an overall $\mathcal{O}(m^2d+md^2)$, where $d$ is the size of the hidden features that are also used in deep competitors for a fair theoretical comparison. The RNN yields the complexity of $\mathcal{O}(md^2)$ after $m$ times sequential operations, while a CNN with the kernel width $k$ increases it considerably to $\mathcal{O}(kmd^2)$. The GCN (i.e., miniGCN) requires $\mathcal{O}(bmd+b^2m)$ owing to its batch-wise graph sampling, where $b$ denotes the size of mini-batches.

\subsection{Model Analysis}
\subsubsection{Ablation Study}
We investigate the performance gain of the proposed SpectralFormer in terms of classification accuracies by adding (in stepwise fashion) different modules (i.e., GSE and CAF) in networks. For that, we conduct extensive ablation experiments on the Indian Pines dataset to verify the effectiveness of these components (or modules) in SpectralFormer for HS image classification applications, as listed in Table \ref{tab:ablation}.

In detail, the classic transformers (ViT) without GSE and CAF modules yield the lowest classification accuracies, which to some extent indicates that the ViT architecture might not be a good fit for the HS image classification. By plugging either GSE or CAF into ViT, the classification results of pixel-wise SpectralFormer are better than that of ViT (beyond around $4\%$ and $3\%$ OAs, respectively). What is better still, the joint exploitation of GSE and CAF can further bring a dramatic performance improvement (more than $4\%$ OA). More remarkably, our SpectralFormer is also capable of capturing the locally spatial semantics of HS images by simply unfolding the patch-wise input. As a result, the patch-wise SpectralFormer performs observably better than pixel-wise ones at an increase at least $3\%$ OA (compared to the second record, i.e., $78.55\%$). 

Interestingly, there is a noteworthy trend in Table \ref{tab:ablation}. That is, the joint use of GSE and CAF tends to obtain the best performance when only using a smaller number of neighboring bands in GSE, compared to that of only using GSE. This might be well explained by that after adding CAF, the spectral information is capable of being learned more efficiently and easier. In other words, the less overlap between spectral bands (i.e., the number of neighboring bands is smaller in GSE) could be enough to obtain a better classification performance, after the CAF module is activated.

\begin{table*}[!t]
\centering
\caption{Quantitative performance of different classification methods in terms of OA, AA, and $\kappa$ as well as the accuracies for each class on the Indian Pines dataset. The best one is shown in bold.}
\resizebox{1\textwidth}{!}{
\begin{tabular}{c||ccc|cccc|c||cc}
\toprule[1.5pt] \multirow{2}{*}{Class No.} & \multicolumn{3}{c|}{Conventional Classifiers} & \multicolumn{4}{c|}{Classic Backbone Networks} & \multirow{2}{*}{Transformers (ViT)} & \multicolumn{2}{c}{SpectralFormer}\\
\cline{2-8} \cline{10-11} & KNN & RF & SVM & 1-D CNN & 2-D CNN & RNN & miniGCN & & pixel-wise & patch-wise\\
\hline \hline
1 & 45.45 & 57.80 & 67.34 & 47.83 & 65.90 & 69.00 & \bf 72.54 & 53.25 & 62.64 & 70.52\\
2 & 46.94 & 56.51 & 67.86 & 42.35 & 76.66 & 58.93 & 55.99 & 66.20 & 66.20 & \bf 81.89\\
3 & 77.72 & 80.98 & \bf 93.48 & 60.87 & 92.39 & 77.17 & 92.93 & 86.41 & 88.59 & 91.30\\
4 & 84.56 & 85.68 & 94.63 & 89.49 & 93.96 & 82.33 & 92.62& 89.71 & 90.16 & \bf 95.53\\
5 & 80.06 & 79.34 & 88.52 & 92.40 & 87.23 & 67.72 & \bf 94.98 & 87.66 & 89.24 & 85.51\\
6 & 97.49 & 95.44 & 94.76 & 97.04 & 97.27 & 89.07 & 98.63 & 89.98 & 95.90 & \bf 99.32\\
7 & 64.81 & 77.56 & 73.86 & 59.69 & 77.23 & 69.06 & 64.71 & 72.22 & \bf 85.19 & 81.81\\
8 & 48.68 & 58.85 & 52.07 & 65.38 & 57.03 & 63.56 & 68.78 & 66.00 & 74.48 & \bf 75.48\\
9 & 44.33 & 62.23 & 72.70 & \bf 93.44 & 72.87 & 65.07 & 69.33 & 57.09 & 72.34 & 73.76\\
10 & 96.30 & 95.06 & 98.77 & 99.38 & \bf 100.00 & 95.06 & 98.77 & 97.53 & 98.15 & 98.77\\
11 & 74.28 & 88.75 & 86.17 & 84.00 & 92.85 & 88.67 & 87.78 & 87.62 & 93.01 & \bf 93.17\\
12 & 15.45 & 54.24 & 71.82 & 86.06 & \bf 88.18 & 50.00 & 50.00 & 63.94 & 60.91 & 78.48\\
13 & 91.11 & 97.78 & 95.56 & 91.11 & \bf 100.00 & 97.78 & \bf 100.00 & 95.56 & \bf 100.00 & \bf 100.00\\
14 & 33.33 & 56.41 & 82.05 & 84.62 & 84.62 & 66.67 & 48.72 & 79.49 & \bf 87.18 & 79.49 \\
15 & 81.82 & 81.82 & 90.91 & \bf 100.00 & \bf 100.00 & 81.82 & 72.73 & 90.91 & 90.91 & \bf 100.00\\
16 & 40.00 & \bf 100.00 & \bf 100.00 & 80.00 & \bf 100.00 & \bf 100.00 & 80.00 & 80.00 & \bf 100.00 & \bf 100.00\\
\hline \hline
OA (\%) & 59.17 & 69.80 & 72.36 & 70.43 & 75.89 & 70.66 & 75.11 & 71.86  & 78.55 & \bf 81.76\\
AA (\%) & 63.90 & 76.78 & 83.16 & 79.60 & 86.64 & 76.37 & 78.03 & 78.97 & 84.68 & \bf 87.81\\
$\kappa$ & 0.5395 & 0.6591 & 0.6888 & 0.6642 & 0.7281 & 0.6673 & 0.7164 & 0.6804 & 0.7554 & \bf 0.7919\\
\bottomrule[1.5pt]
\end{tabular}}
\label{tab:IP}
\end{table*}

\begin{table*}[!t]
\centering
\caption{Quantitative performance of different classification methods in terms of OA, AA, and $\kappa$, as well as the accuracies for each class on the Pavia University dataset. The best one is shown in bold.}
\resizebox{1\textwidth}{!}{
\begin{tabular}{c||ccc|cccc|c||cc}
\toprule[1.5pt] \multirow{2}{*}{Class No.} & \multicolumn{3}{c|}{Conventional Classifiers} & \multicolumn{4}{c|}{Classic Backbone Networks} & \multirow{2}{*}{Transformers (ViT)} & \multicolumn{2}{c}{SpectralFormer}\\
\cline{2-8} \cline{10-11} & KNN & RF & SVM & 1-D CNN & 2-D CNN & RNN & miniGCN & & pixel-wise & patch-wise\\
\hline \hline
1 & 73.86 & 79.81 & 74.22 & 88.90 & 80.98 & 84.01 & \bf 96.35 & 71.51 & 82.95 & 82.73\\
2 & 64.31 & 54.90 & 52.79 & 58.81 & 81.70 & 66.95 & 89.43 & 76.82 & \bf 95.23 & 94.03\\
3 & 55.10 & 46.34 & 65.45 & 73.11 & 67.99 & 58.46 & \bf 87.01 & 46.39 & 78.18 & 73.66\\
4 & 94.95 & \bf 98.73 & 97.42 & 82.07 & 97.36 & 97.70 & 94.26 & 96.39 & 87.95 & 93.75\\
5 & 99.19 & 99.01 & 99.46 & 99.46 & 99.64 & 99.10 & \bf 99.82 & 99.19 & 99.46 & 99.28\\
6 & 65.16 & 75.94 & 93.48 & \bf 97.92 & 97.59 & 83.18 & 43.12 & 83.18 & 65.84 & 90.75\\
7 & 84.30 & 78.70 & 87.87 & 88.07 & 82.47 & 83.08 & \bf 90.96 & 83.08 & 92.35 & 87.56\\
8 & 84.10 & 90.22 & 89.39 & 88.14 & \bf 97.62 & 89.63 & 77.42 & 89.63 & 85.26 & 95.81\\
9 & 98.36 & 97.99 & 99.87 & 99.87 & 95.60 & 96.48 & 87.27 & 96.48 & \bf 100.00 & 94.21\\
\hline \hline
OA (\%) & 70.53 & 69.67 & 70.82 & 75.50 & 86.05 & 77.13 & 79.79 & 76.99 & 87.94 & \bf 91.07\\
AA (\%) & 79.68 & 80.18 & 84.44 & 86.26 & 88.99 & 84.29 & 85.07 & 80.22 & 87.47 & \bf 90.20\\
$\kappa$ & 0.6268 & 0.6237 & 0.6423 & 0.6948 & 0.8187 & 0.7101 & 0.7367 & 0.7010 & 0.8358 & \bf 0.8805\\
\bottomrule[1.5pt]
\end{tabular}}
\label{tab:PU}
\end{table*}

\subsubsection{Parameter Sensitivity Analysis}
Apart from the learnable parameters in networks and hyper-parameters required in the training process, the number of neighboring bands in GES plays a vital role in the final classification performance. It is, therefore, indispensable to explore the proper parameter range. Similarly, we investigate the parameter sensitivity on the Indian pines dataset in an ablation manner, i.e., only with GSE and the joint use of GSE and CAF. Table \ref{tab:Sensitivity} lists the changing trend of the classification accuracies with the gradual increase of the number of grouped bands in terms of OA, AA, and $\kappa$. A common conclusion is that GSE can better excavate subtle spectral discrepancies by effectively capturing locally spectral embeddings from neighboring bands. Within a certain range, the parameter is not sensitive to the classification performance. This provides great potentials of the proposed model in the practical applications. In other words, the parameter can be simply and directly used for other datasets.

We also make a quantitative comparison among short-range SC (e.g., in ResNet), long-range SC (e.g., in U-Net), and middle-range (i.e., our CAF module) SC in transformers, in order to verify the effectiveness of the proposed SpectralFormer in processing spectral data. Table \ref{tab:skipconnection} quantifies the classification performance comparison of using short-, middle-, and long-range SCs, respectively, on the Indian Pines dataset. In general, the ViT with long-range SC yields a poor performance, possibly since such an SC strategy might fail to sufficiently fuse and convey long-range cross-layer features, tending to lose partially ``important'' information. This demonstrates the superiority of the CAF module that can exchange information across different layers more effectively (\textit{cf.} short-range SC) and reduce the information loss (\textit{cf.} long-range SC).

\begin{figure}[!t]
	  \centering
			\includegraphics[width=0.45\textwidth]{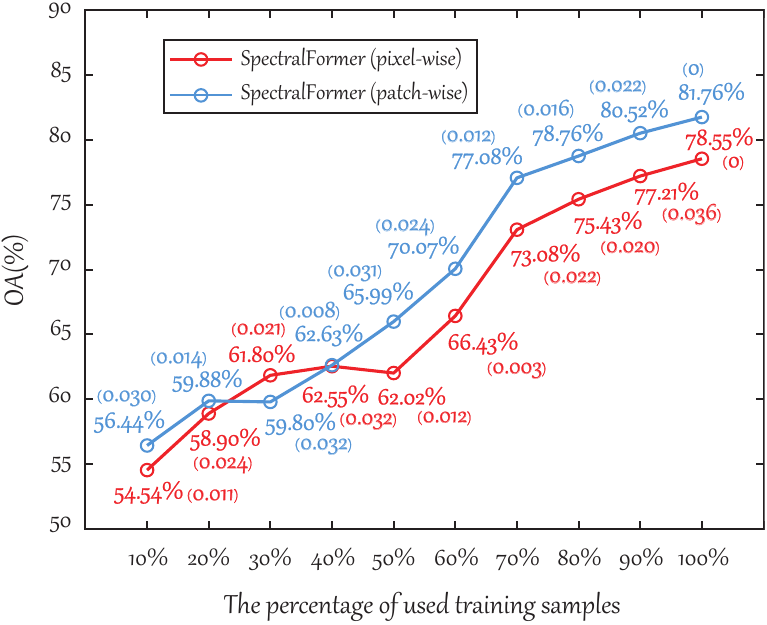}
        \caption{Classification results (OA) obtained by our proposed SpectralFormer (pixel-wise and patch-wise) with the varying number of used training samples on the Indian Pines dataset.}
\label{fig:trainingsize}
\end{figure}

Furthermore, we randomly selected varying number of training samples from the given training set on the Indian Pines dataset out of $10$ runs in the proportion of [$10\%$, $20\%$, ... , $100\%$] at intervals of $10\%$. The average results with standard deviation values in terms of the OA obtained by the proposed SpectralFormer (including pixel-wise and patch-wise versions) are reported in Fig. \ref{fig:trainingsize}. There is a basically reasonable trend in OA's results (see Fig. \ref{fig:trainingsize}). That is, with the increase of the percentage of used training samples, the classification performance gradually improves. Note that the OAs are tending towards stability when more training samples (e.g., $80\%$, $90\%$, and $100\%$) are involved, showing the stability of the proposed SpectralFormer to a great extent. Moreover, the patch-wise SpectralFormer observably outperforms the pixel-wise one, as expected.

\begin{table*}[!t]
\centering
\caption{Quantitative performance of different classification methods in terms of OA, AA, and $\kappa$ as well as the accuracies for each class on the Houston2013 dataset. The best one is shown in bold.}
\resizebox{1\textwidth}{!}{
\begin{tabular}{c||ccc|cccc|c||cc}
\toprule[1.5pt] \multirow{2}{*}{Class No.} & \multicolumn{3}{c|}{Conventional Classifiers} & \multicolumn{4}{c|}{Classic Backbone Networks} & \multirow{2}{*}{Transformers (ViT)} & \multicolumn{2}{c}{SpectralFormer}\\
\cline{2-8} \cline{10-11} & KNN & RF & SVM & 1-D CNN & 2-D CNN & RNN & miniGCN & & pixel-wise & patch-wise\\
\hline \hline
1 & 83.19 & 83.38 & 83.00 & 87.27 & 85.09 & 82.34 & \bf 98.39 & 82.81 & 83.48 & 81.86\\
2 & 95.68 & 98.40 & 98.40 & 98.21 & 99.91 & 94.27 & 92.11 & 96.62 & 95.58 & \bf 100.00\\
3 & 99.41 & 98.02 & 99.60 & \bf 100.00 & 77.23 & 99.60 & 99.60 & 99.80 & 99.60 & 95.25\\
4 & 97.92 & 97.54 & 98.48 & 92.99 & 97.73 & 97.54 & 96.78 & \bf 99.24 & 99.15 & 96.12\\
5 & 96.12 & 96.40 & 97.82 & 97.35 & \bf 99.53 & 93.28 & 97.73 & 97.73 & 97.44 & \bf 99.53\\
6 & 92.31 & \bf 97.20 & 90.91 & 95.10 & 92.31 & 95.10 & 95.10 & 95.10 & 95.10 & 94.41\\
7 & 80.88 & 82.09 & 90.39 & 77.33 & \bf 92.16 & 83.77 & 57.28 & 76.77 & 88.99 & 83.12\\
8 & 48.62 & 40.65 & 40.46 & 51.38 & \bf 79.39 & 56.03 & 68.09 & 55.65 & 73.31 & 76.73\\
9 & 72.05 & 69.78 & 41.93 & 27.95 & \bf 86.31 & 72.14 & 53.92 & 67.42 & 71.86 & 79.32\\
10 & 53.19 & 57.63 & 62.64 & \bf 90.83 & 43.73 & 84.17 & 77.41 & 68.05 & 87.93 & 78.86\\
11 & 86.24 & 76.09 & 75.43 & 79.32 & 87.00 & 82.83 & 84.91 & 82.35 & 80.36 & \bf 88.71\\
12 & 44.48 & 49.38 & 60.04 & 76.56 & 66.28 & 70.61 & 77.23 & 58.50 & 70.70 & \bf 87.32\\
13 & 28.42 & 61.40 & 49.47 & 69.47 & \bf 90.18 & 69.12 & 50.88 & 60.00 & 71.23 & 72.63\\
14 & 97.57 & \bf 99.60 & 98.79 & 99.19 & 90.69 & 98.79 & 98.38 & 98.79 & 98.79 & \bf 100.00\\
15 & 98.10 & 97.67 & 97.46 & 98.10 & 77.80 & 95.98 & 98.52 & 98.73 & 98.73 & \bf 99.79\\
\hline \hline
OA (\%) & 77.30 & 77.48 & 76.91 & 80.04 & 83.72 & 83.23 & 81.71 & 80.41 & 86.14 & \bf 88.01\\
AA (\%) & 78.28 & 80.35 & 78.99 & 82.74 & 84.35 & 85.04 & 83.09 & 82.50 & 87.48 & \bf 88.91\\
$\kappa$ & 0.7538 & 0.7564 & 74.94 & 0.7835 & 0.8231 & 0.8183 & 0.8018 & 0.7876 & 0.8497 & \bf 0.8699\\
\bottomrule[1.5pt]
\end{tabular}}
\label{tab:HH}
\end{table*}

\begin{figure*}[!t]
	  \centering
			\includegraphics[width=0.9\textwidth]{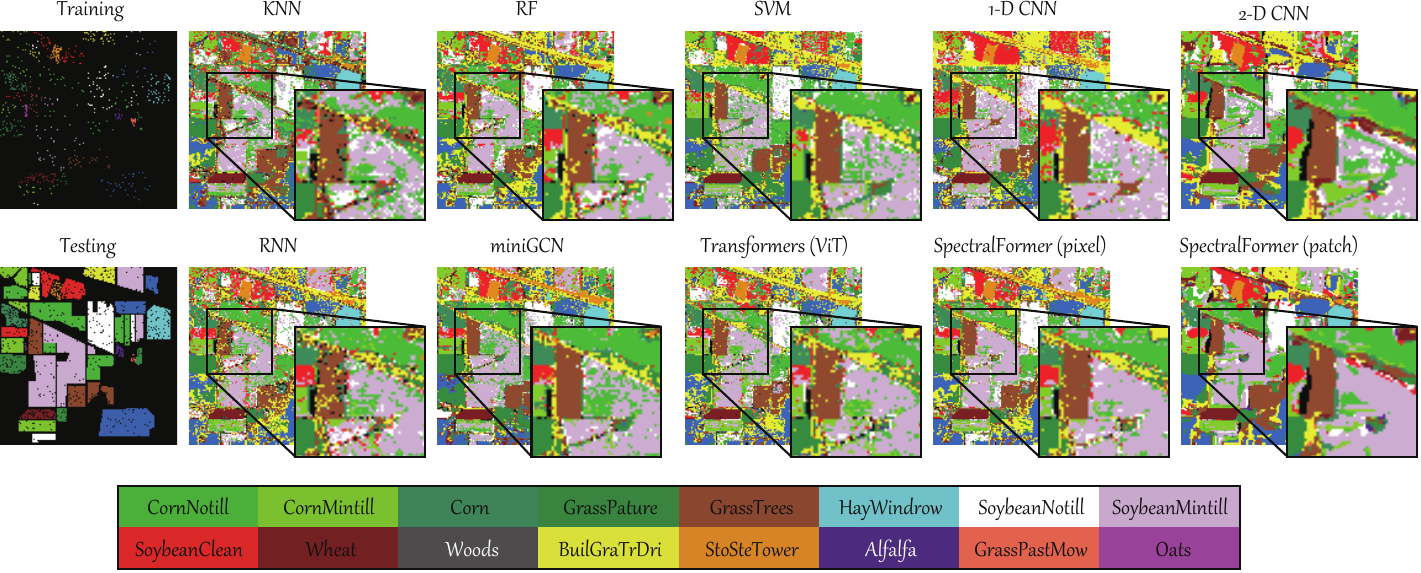}
        \caption{Spatial distribution of training and testing sets, and the classification maps obtained by different models on the Indian Pines dataset, where a ROI zoomed in $2$ times is highlighted for more detailed observation.}
\label{fig:CM_IP}
\end{figure*}

\begin{figure*}[!t]
	  \centering
			\includegraphics[width=1\textwidth]{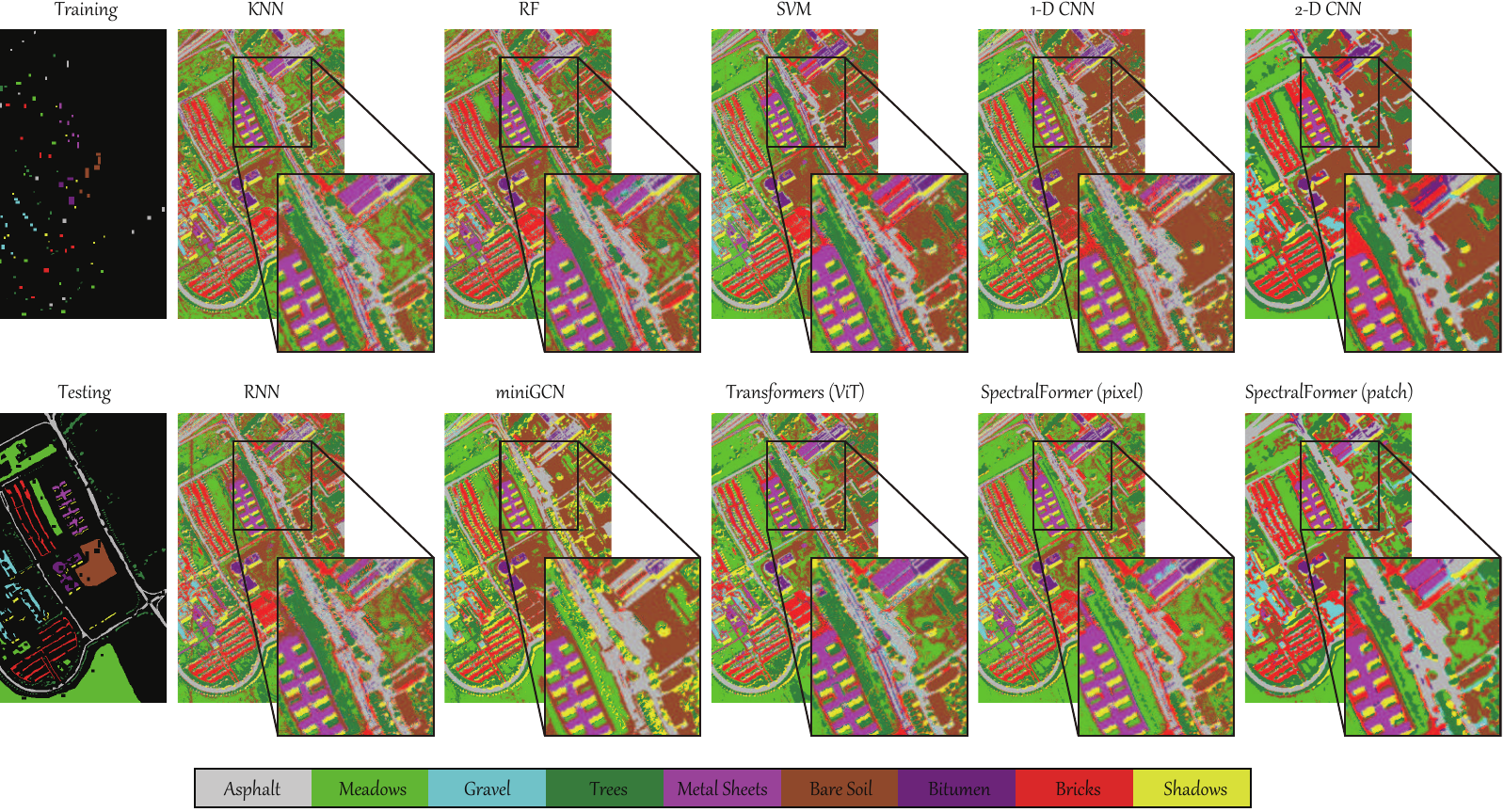}
        \caption{Spatial distribution of training and testing sets, and the classification maps obtained by different models on the Pavia University dataset, where a ROI zoomed in $2$ times is highlighted for more detailed observation.}
\label{fig:CM_PU}
\end{figure*}

\begin{figure*}[!t]
	  \centering
			\includegraphics[width=0.9\textwidth]{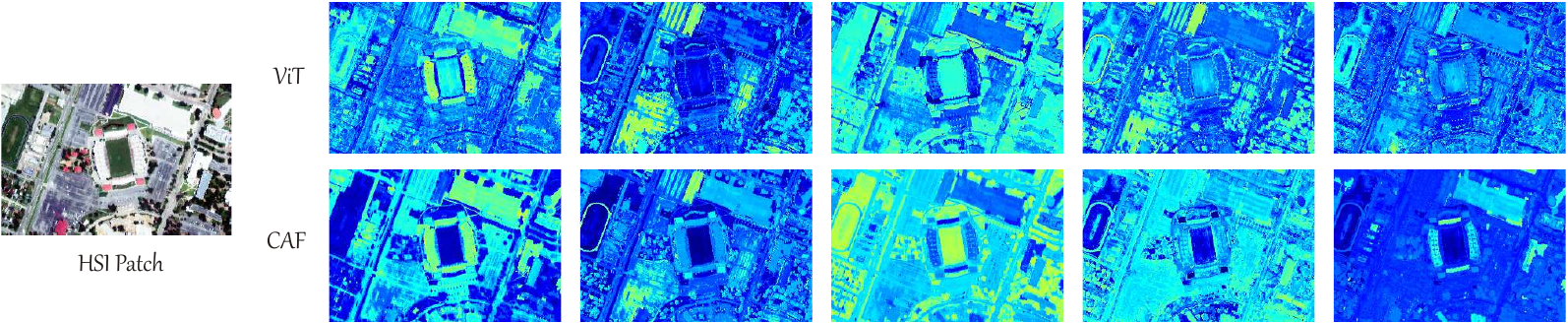}
        \caption{Visualization of selected encoder output features obtained with only CAF and without CAF (i.e., the original ViT).}
\label{fig:vis_feature}
\end{figure*}

\begin{figure*}[!t]
	  \centering
			\includegraphics[width=1\textwidth]{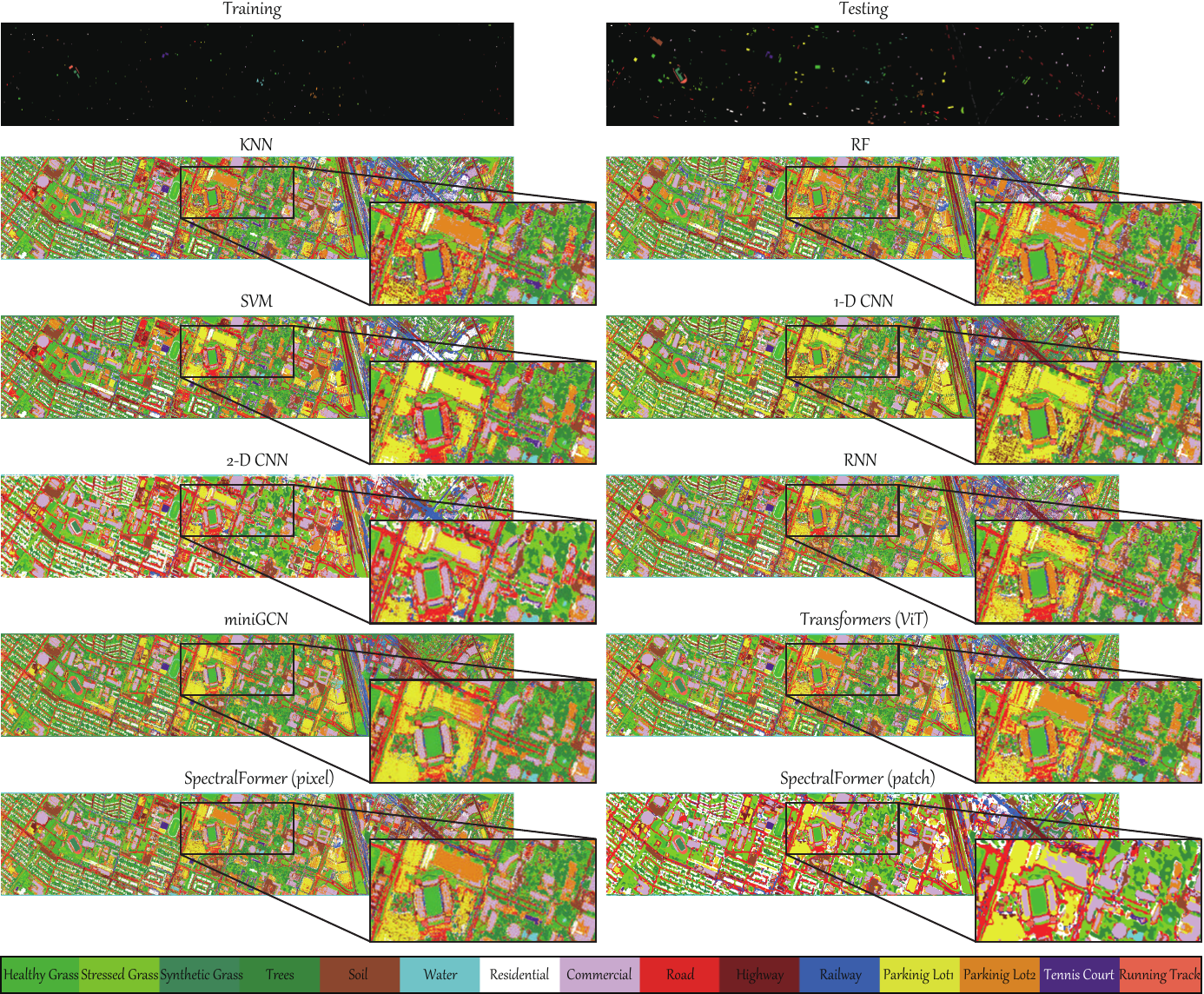}
        \caption{Spatial distribution of training and testing sets, and the classification maps obtained by different models on the Houston2013 dataset, where a ROI zoomed in $2$ times is highlighted for more detailed observation.}
\label{fig:CM_HH}
\end{figure*}

\subsection{Quantitative Results and Analysis}
Quantitative classification results in terms of three overall indices, i.e., OA, AA, and $\kappa$, and the accuracies for each class are reported in Tables \ref{tab:IP}, \ref{tab:PU}, and \ref{tab:HH} for Indian Pines, Pavia University, and Houston2013 HS datasets, respectively.

Overall, the conventional classifiers, e.g., KNN, RF, SVM, achieve similar classification performance on all three data sets, except the accuracies in terms of OA, AA, and $\kappa$ with KNN on the Indian Pines dataset (which are far inferior to those using RF and SVM). Owing to the powerful learning ability of DL techniques, classic backbone networks, e.g., 1-D CNN, 2-D CNN, RNN, miniGCN, observably perform better than aforementioned conventional classifiers, i.e., KNN, RF, SVM. The results, to a great extent, demonstrate the value and practicality of DL-based approaches in HS image classification. Without any convolutional and recurrent operations, transformers extract finer spectral representations from the sequence perspectives, yielding a comparable performance to CNNs-, RNNs-, or GCNs-based models. 

Although transformers are capable of capturing globally sequential information, the ability in modeling some key factors -- locally spectral discrepancies -- remains limited, leads to a performance bottleneck. To overcome this problem, the proposed SpectralFormer fully extracts local spectral information from neighboring bands, dramatically improving the classification performance. In particular, the pixel-wise SpectralFormer unexpectedly outperforms others, even though comparing to CNNs-based methods considering spatial contents. Undoubtedly, the patch-wise SpectralFormer obtains higher classification accuracies which are far superior to those obtained by other competitors, since the spatial-contextual information is jointly considered in the sequential feature extraction process.

In addition, for those challenging classes that have limited training samples (e.g., \textit{Grass Pasture Mowed}, \textit{Oats}) and imbalanced (or noisy) samples (e.g., \textit{Corn Mintill}, \textit{Grass Pasture}, \textit{Hay Windrowed}) on the Indian Pines data, SpectralFormer, either pixel-wise or patch-wise input, tends to obtain better classification performance by focusing on particular absorption positions of spectral profiles. By contrary, despite the excellent representation ability for spectral sequence data, transformers (ViT) fail to accurately capture the detailed spectral absorption or change due to the weak modeling ability in locally spectral discrepancies.

\subsection{Visual Evaluation}
We make a qualitative evaluation by visualizing the classification maps obtained by different methods. Figs. \ref{fig:CM_IP}, \ref{fig:CM_PU}, and \ref{fig:CM_HH} provide the obtained results for the Indian Pines, Pavia University, and Houston2013 datasets, respectively. Roughly, conventional classification models (e.g., KNN, RF, SVM) tend to generate salt and pepper noises in classification maps of three considered datasets. This indirectly indicates that these classifiers fail to accurately identifying the materials of objects. Not surprisingly, DL-based models, e.g., CNNs, RNNs, GCNs, obtain relatively smooth classification maps, owing to their powerful nonlinear data fitting ability. As an emerging network architecture, transformers (ViT used in our case) can extract highly sequential representations from HS images, leading to visualized classification maps that are comparable to the above classic backbone networks. By enhancing spectrally neighboring information and conveying ``memory'' information across layers more effectively, our proposed SpectralFormer obtains highly desirable classification maps, especially in terms of texture and edge details. Furthermore, we selected a region of interest (ROI) (from Figs. \ref{fig:CM_IP}, \ref{fig:CM_PU}, and \ref{fig:CM_HH}) zoomed-in $2$ times to highlight the differences in classification maps between different models, further evaluating their classification performance more intuitively. As can be seen from these ROIs, a remarkable phenomenon is that our methods, i.e., pixel-wise and patch-wise SpectralFormers, show more realistic and finer details. In particular, the results of our methods have less noisy points compared to those pixel-wise methods, e.g., KNN, RF, SVM, 1-D CNN, RNN, miniGCN, and ViT, but also avoid over-smoothness in edges or some small semantic objects (\textit{cf.} 2-D CNN), which yields more accurate classification performance.

\noindent \textbf{Feature Visualization.} Fig. \ref{fig:vis_feature} visualizes selected encoder output features using the proposed SpectralFormer framework with only CAF and without CAF (i.e., ViT). We selectively pick out some representative feature maps for visual comparison, where the visualization results of using the CAF module have finer appearance (e.g., the edge or outline of objects, textural structure, etc.) than those without CAF. This also demonstrates the effectiveness and superiority of the designed CAF module from the visual perspective. 

\section{Conclusion}
HS images are typically collected (or represented) as a data cube with spatial-spectral information, which can be generally regarded as a sequence of data along the spectral dimension. Unlike CNNs, that focus mainly on contextual information modeling, transformers have been proven to be a powerful architecture in characterizing the sequential properties globally. However, the classic transformers-based vision networks, e.g., ViT, inevitably suffer from performance degradation when processing HS-like data. This might be explained well by the fact that ViT fails to model locally detailed spectral discrepancies and convey ``memory''-like components (from shallow to deep layers) effectively. To this end, in this paper we propose a new transformers-based backbone network, called SpectralFormer, which is more focused on extracting spectral information. Without using any convolution or recurrent units, the proposed SpectralFormer can achieve state-of-the-art classification results for HS images.

In the future, we will investigate strategies to further improve the transformers-based architecture by utilizing more advanced techniques, e.g., attention, self-supervised learning, making it more applicable to the HS image classification task, and also attempt to establish a lightweight transformers-based network to reduce the network complexity while maintaining its performance. Moreover, we would also like to embed more physical characteristics of spectral bands and prior knowledge of HS images into the proposed framework, yielding more interpretable deep models. Furthermore, the number of skipped and connected encoders in the CAF module is an important factor that might be capable of improving the classification performance of the proposed SpectralFormer, which should be paid more attention to in future work.

\bibliographystyle{ieeetr}
\bibliography{HDF_ref}

\begin{thebibliography}{10}

\bibitem{hong2021interpretable}
D.~Hong, W.~He, N.~Yokoya, J.~Yao, L.~Gao, L.~Zhang, J.~Chanussot, and X.~X.
  Zhu, ``Interpretable hyperspectral artificial intelligence: When non-convex
  modeling meets hyperspectral remote sensing,'' {\em IEEE Geosci. Remote Sens.
  Mag.}, vol.~9, no.~2, pp.~52--87, 2021.

\bibitem{wang2017hyperspectral}
Y.~Wang, J.~Peng, Q.~Zhao, Y.~Leung, X.-L. Zhao, and D.~Meng, ``Hyperspectral
  image restoration via total variation regularized low-rank tensor
  decomposition,'' {\em IEEE J. Sel. Top. Appl. Earth Obs. Remote Sens.},
  vol.~11, no.~4, pp.~1227--1243, 2017.

\bibitem{cao2018robust}
W.~Cao, K.~Wang, G.~Han, J.~Yao, and A.~Cichocki, ``A robust pca approach with
  noise structure learning and spatial--spectral low-rank modeling for
  hyperspectral image restoration,'' {\em IEEE J. Sel. Top. Appl. Earth Obs.
  Remote Sens.}, vol.~11, no.~10, pp.~3863--3879, 2018.

\bibitem{wang2021}
M.~Wang, Q.~Wang, J.~Chanussot, and D.~Hong, ``$l_{0}$-$l_{1}$ hybrid total
  variation regularization and its applications on hyperspectral image mixed
  noise removal and compressed sensing,'' {\em IEEE Trans. Geosci. Remote
  Sens.}, 2021.
\newblock DOI: 10.1109/TGRS.2021.3055516.

\bibitem{peng2021low}
J.~Peng, W.~Sun, H.-C. Li, W.~Li, X.~Meng, C.~Ge, and Q.~Du, ``Low-rank and
  sparse representation for hyperspectral image processing: A review,'' {\em
  IEEE Geosci. Remote Sens. Mag.}, 2021.

\bibitem{hong2021joint}
D.~Hong, N.~Yokoya, J.~Chanussot, J.~Xu, and X.~X. Zhu, ``Joint and progressive
  subspace analysis (jpsa) with spatial-spectral manifold alignment for
  semi-supervised hyperspectral dimensionality reduction,'' {\em IEEE Trans.
  Cybern.}, vol.~51, no.~7, pp.~3602--3615, 2021.

\bibitem{luo2020semisupervised}
F.~Luo, T.~Guo, Z.~Lin, J.~Ren, and X.~Zhou, ``Semisupervised hypergraph
  discriminant learning for dimensionality reduction of hyperspectral image,''
  {\em IEEE J. Sel. Top. Appl. Earth Obs. Remote Sens.}, vol.~13,
  pp.~4242--4256, 2020.

\bibitem{yao2019nonconvex}
J.~Yao, D.~Meng, Q.~Zhao, W.~Cao, and Z.~Xu, ``Nonconvex-sparsity and
  nonlocal-smoothness-based blind hyperspectral unmixing,'' {\em IEEE Trans.
  Image Process.}, vol.~28, no.~6, pp.~2991--3006, 2019.

\bibitem{hong2019augmented}
D.~Hong, N.~Yokoya, J.~Chanussot, and X.~Zhu, ``An augmented linear mixing
  model to address spectral variability for hyperspectral unmixing,'' {\em IEEE
  Trans. Image Process.}, vol.~28, no.~4, pp.~1923--1938, 2019.

\bibitem{yuan2020improved}
Y.~Yuan, Z.~Zhang, and Q.~Wang, ``Improved collaborative non-negative matrix
  factorization and total variation for hyperspectral unmixing,'' {\em IEEE J.
  Sel. Top. Appl. Earth Obs. Remote Sens.}, vol.~13, pp.~998--1010, 2020.

\bibitem{gao2021cycu}
L.~Gao, Z.~Han, D.~Hong, B.~Zhang, and J.~Chanussot, ``Cycu-net:
  Cycle-consistency unmixing network by learning cascaded autoencoders,'' {\em
  IEEE Trans. Geosci. Remote Sens.}, 2021.
\newblock DOI: 10.1109/TGRS.2021.3064958.

\bibitem{hong2021endmember}
D.~Hong, L.~Gao, J.~Yao, N.~Yokoya, J.~Chanussot, U.~Heiden, and B.~Zhang,
  ``Endmember-guided unmixing network (egu-net): A general deep learning
  framework for self-supervised hyperspectral unmixing,'' {\em IEEE Trans.
  Neural Netw. Learn. Syst.}, May 2021.
\newblock DOI: 10.1109/TNNLS.2021.3082289.

\bibitem{hong2019learning}
D.~Hong, N.~Yokoya, J.~Chanussot, J.~Xu, and X.~Zhu, ``Learning to propagate
  labels on graphs: An iterative multitask regression framework for
  semi-supervised hyperspectral dimensionality reduction,'' {\em ISPRS J.
  Photogramm. Remote Sens.}, vol.~158, pp.~35--49, 2019.

\bibitem{peng2018self}
J.~Peng, W.~Sun, and Q.~Du, ``Self-paced joint sparse representation for the
  classification of hyperspectral images,'' {\em IEEE Trans. Geosci. Remote
  Sens.}, vol.~57, no.~2, pp.~1183--1194, 2018.

\bibitem{hong2020invariant}
D.~Hong, X.~Wu, P.~Ghamisi, J.~Chanussot, N.~Yokoya, and X.~X. Zhu, ``Invariant
  attribute profiles: A spatial-frequency joint feature extractor for
  hyperspectral image classification,'' {\em IEEE Trans. Geosci. Remote Sens.},
  vol.~58, no.~6, pp.~3791--3808, 2020.

\bibitem{li2020ensemble}
Q.~Li, B.~Zheng, B.~Tu, J.~Wang, and C.~Zhou, ``Ensemble emd-based
  spectral-spatial feature extraction for hyperspectral image classification,''
  {\em IEEE J. Sel. Top. Appl. Earth Obs. Remote Sens.}, vol.~13,
  pp.~5134--5148, 2020.

\bibitem{rasti2020feature}
B.~Rasti, D.~Hong, R.~Hang, P.~Ghamisi, X.~Kang, J.~Chanussot, and
  J.~Benediktsson, ``Feature extraction for hyperspectral imagery: The
  evolution from shallow to deep: Overview and toolbox,'' {\em IEEE Geosci.
  Remote Sens. Mag.}, vol.~8, no.~4, pp.~60--88, 2020.

\bibitem{lecun2015deep}
Y.~LeCun, Y.~Bengio, and G.~Hinton, ``Deep learning,'' {\em Nature}, vol.~521,
  no.~7553, pp.~436--444, 2015.

\bibitem{zhao2020joint}
X.~Zhao, R.~Tao, W.~Li, H.-C. Li, Q.~Du, W.~Liao, and W.~Philips, ``Joint
  classification of hyperspectral and lidar data using hierarchical random walk
  and deep cnn architecture,'' {\em IEEE Trans. Geosci. Remote Sens.}, vol.~58,
  no.~10, pp.~7355--7370, 2020.

\bibitem{zhang2018feature}
M.~Zhang, W.~Li, Q.~Du, L.~Gao, and B.~Zhang, ``Feature extraction for
  classification of hyperspectral and lidar data using patch-to-patch cnn,''
  {\em IEEE Trans. Cybern.}, vol.~50, no.~1, pp.~100--111, 2018.

\bibitem{chen2014deep}
Y.~Chen, Z.~Lin, X.~Zhao, G.~Wang, and Y.~Gu, ``Deep learning-based
  classification of hyperspectral data,'' {\em IEEE J. Sel. Top. Appl. Earth
  Obs. Remote Sens.}, vol.~7, no.~6, pp.~2094--2107, 2014.

\bibitem{chen2016deep}
Y.~Chen, H.~Jiang, C.~Li, X.~Jia, and P.~Ghamisi, ``Deep feature extraction and
  classification of hyperspectral images based on convolutional neural
  networks,'' {\em IEEE Trans. Geosci. Remote Sens.}, vol.~54, no.~10,
  pp.~6232--6251, 2016.

\bibitem{hang2019cascaded}
R.~Hang, Q.~Liu, D.~Hong, and P.~Ghamisi, ``Cascaded recurrent neural networks
  for hyperspectral image classification,'' {\em IEEE Trans. Geosci. Remote
  Sens.}, vol.~57, no.~8, pp.~5384--5394, 2019.

\bibitem{zhu2018generative}
L.~Zhu, Y.~Chen, P.~Ghamisi, and J.~A. Benediktsson, ``Generative adversarial
  networks for hyperspectral image classification,'' {\em IEEE Trans. Geosci.
  Remote Sens.}, vol.~56, no.~9, pp.~5046--5063, 2018.

\bibitem{paoletti2018capsule}
M.~E. Paoletti, J.~M. Haut, R.~Fernandez-Beltran, J.~Plaza, A.~Plaza, J.~Li,
  and F.~Pla, ``Capsule networks for hyperspectral image classification,'' {\em
  IEEE Trans. Geosci. Remote Sens.}, vol.~57, no.~4, pp.~2145--2160, 2018.

\bibitem{hong2021graph}
D.~Hong, L.~Gao, J.~Yao, B.~Zhang, A.~Plaza, and J.~Chanussot, ``Graph
  convolutional networks for hyperspectral image classification,'' {\em IEEE
  Trans. Geosci. Remote Sens.}, vol.~59, no.~7, pp.~5966--5978, 2021.

\bibitem{vaswani2017attention}
A.~Vaswani, N.~Shazeer, N.~Parmar, J.~Uszkoreit, L.~Jones, A.~N. Gomez,
  L.~Kaiser, and I.~Polosukhin, ``Attention is all you need,'' {\em arXiv
  preprint arXiv:1706.03762}, 2017.

\bibitem{li2019deep}
S.~Li, W.~Song, L.~Fang, Y.~Chen, P.~Ghamisi, and J.~A. Benediktsson, ``Deep
  learning for hyperspectral image classification: An overview,'' {\em IEEE
  Trans. Geosci. Remote Sens.}, vol.~57, no.~9, pp.~6690--6709, 2019.

\bibitem{abdi2010principal}
H.~Abdi and L.~J. Williams, ``Principal component analysis,'' {\em Wiley
  Interdiscip. Rev. Comput. Stat.}, vol.~2, no.~4, pp.~433--459, 2010.

\bibitem{bengio1994learning}
Y.~Bengio, P.~Simard, and P.~Frasconi, ``Learning long-term dependencies with
  gradient descent is difficult,'' {\em IEEE Trans. Neural Netw.}, vol.~5,
  no.~2, pp.~157--166, 1994.

\bibitem{ke2020rethinking}
G.~Ke, D.~He, and T.-Y. Liu, ``Rethinking the positional encoding in language
  pre-training,'' {\em arXiv preprint arXiv:2006.15595}, 2020.

\bibitem{dong2021attention}
Y.~Dong, J.-B. Cordonnier, and A.~Loukas, ``Attention is not all you need: Pure
  attention loses rank doubly exponentially with depth,'' {\em arXiv preprint
  arXiv:2103.03404}, 2021.

\bibitem{dosovitskiy2020image}
A.~Dosovitskiy, L.~Beyer, A.~Kolesnikov, D.~Weissenborn, X.~Zhai,
  T.~Unterthiner, M.~Dehghani, M.~Minderer, G.~Heigold, S.~Gelly, {\em et~al.},
  ``An image is worth 16$\times$16 words: Transformers for image recognition at
  scale,'' {\em arXiv preprint arXiv:2010.11929}, 2020.

\bibitem{wang2018non}
X.~Wang, R.~Girshick, A.~Gupta, and K.~He, ``Non-local neural networks,'' in
  {\em Proc. CVPR}, pp.~7794--7803, 2018.

\bibitem{he2016deep}
K.~He, X.~Zhang, S.~Ren, and J.~Sun, ``Deep residual learning for image
  recognition,'' in {\em Proc. CVPR}, pp.~770--778, 2016.

\bibitem{ronneberger2015u}
O.~Ronneberger, P.~Fischer, and T.~Brox, ``U-net: Convolutional networks for
  biomedical image segmentation,'' in {\em Proc. MICCAI}, pp.~234--241,
  Springer, 2015.

\bibitem{hong2021more}
D.~Hong, L.~Gao, N.~Yokoya, J.~Yao, J.~Chanussot, D.~Qian, and B.~Zhang, ``More
  diverse means better: Multimodal deep learning meets remote-sensing imagery
  classification,'' {\em IEEE Trans. Geosci. Remote Sens.}, vol.~59, no.~5,
  pp.~4340--4354, 2021.

\bibitem{hendrycks2016gaussian}
D.~Hendrycks and K.~Gimpel, ``Gaussian error linear units (gelus),'' {\em arXiv
  preprint arXiv:1606.08415}, 2016.

\bibitem{loshchilov2017decoupled}
I.~Loshchilov and F.~Hutter, ``Decoupled weight decay regularization,'' {\em
  arXiv preprint arXiv:1711.05101}, 2017.

\bibitem{kingma2014adam}
D.~Kingma and J.~Ba, ``Adam: A method for stochastic optimization,'' {\em arXiv
  preprint arXiv:1412.6980}, 2014.

\end{thebibliography}

\end{document}